\documentclass[letterpaper, 10 pt, conference]{ieeeconf}  

\IEEEoverridecommandlockouts                              

\usepackage{amsmath,amssymb,amsfonts}
\usepackage{algorithmic}
\usepackage{graphicx}
\usepackage{textcomp}
\usepackage{xcolor}

\usepackage[style=ieee,dashed=false]{biblatex}

\DeclareSourcemap{
  \maps{
    \map{
      \pertype{article}
      \step[fieldset=language, null]
      \step[fieldset=url, null]
      \step[fieldset=doi, null]
      \step[fieldset=issn, null]
      \step[fieldset=isbn, null]
      \step[fieldset=note, null]
      \step[fieldset=editor, null]
      \step[fieldset=urldate, null]
      \step[fieldset=file, null]
    }
  }
}
\DeclareSourcemap{
  \maps{
    \map{
      \pertype{inproceedings}
      \step[fieldset=language, null]
      \step[fieldset=url, null]
      \step[fieldset=doi, null]
      \step[fieldset=issn, null]
      \step[fieldset=isbn, null]
      \step[fieldset=note, null]
      \step[fieldset=editor, null]
      \step[fieldset=urldate, null]
      \step[fieldset=file, null]
    }
  }
}
\DeclareSourcemap{
  \maps{
    \map{
      \pertype{incollection}
      \step[fieldset=language, null]
      \step[fieldset=url, null]
      \step[fieldset=doi, null]
      \step[fieldset=issn, null]
      \step[fieldset=isbn, null]
      \step[fieldset=note, null]
      \step[fieldset=editor, null]
      \step[fieldset=urldate, null]
      \step[fieldset=file, null]
    }
  }
}

\title{\LARGE \bf
Hovering Control of Flapping Wings in Tandem with Multi-Rotors
}

\author{Aniket Dhole$^{1}$, Bibek Gupta$^{1}$, Adarsh Salagame$^{1}$, Xuejian Niu$^1$, Yizhe Xu$^1$, Kaushik Venkatesh$^{1}$, \\
Paul Ghanem$^{1}$, Ioannis Mandralis$^{2}$, Eric Sihite$^{2}$, and Alireza Ramezani$^{1*}$
\thanks{$^{1}$Authors are with the Silicon Synapse Labs, Department of Electrical and Computer Engineering, Northeastern University, Boston, USA. Emails: 
        {\tt\small dhole.an, gupta.bi, salagame.a, a.ramezani@northeastern.edu}}%
\thanks{$^{2}$Author is with the Department of Aerospace Engineering, California Institute of Technology, Pasadena, USA. Email: 
        {\tt\small esihite@caltech.edu}}%
\thanks{$^{*}$Corresponding author.}%
}

\begin{document}



\maketitle
\thispagestyle{empty}
\pagestyle{empty}

\begin{abstract}
This work briefly covers our efforts to stabilize the flight dynamics of Northeatern's tailless bat-inspired micro aerial vehicle, Aerobat. Flapping robots are not new. A plethora of examples is mainly dominated by insect-style design paradigms that are passively stable. However, Aerobat, in addition for being tailless, possesses morphing wings that add to the inherent complexity of flight control. The robot can dynamically adjust its wing platform configurations during gaitcycles, increasing its efficiency and agility. We employ a guard design with manifold small thrusters to stabilize Aerobat's position and orientation in hovering, a flapping system in tandem with a multi-rotor. For flight control purposes, we take an approach based on assuming the guard cannot observe Aeroat's states. Then, we propose an observer to estimate the unknown states of the guard which are then used for closed-loop hovering control of the Guard-Aerobat platform. 
\end{abstract}


\section{Introduction} 

\begin{figure}
\vspace{0.08in}
    \centering
    \includegraphics[width=1\linewidth]{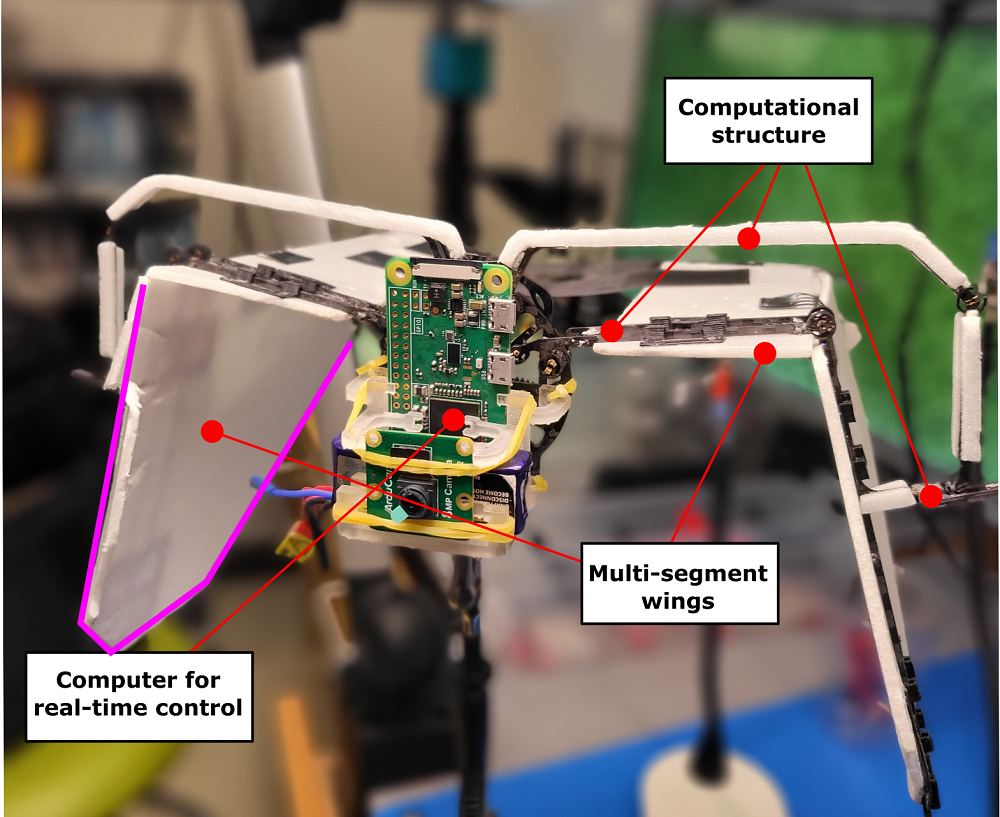}
    \caption{Illustrates Aerobat and its articulated wings that can dynamically reconfigure during a gait cycle.}
    \label{fig:cover}
\vspace{-0.08in}
\end{figure}

This paper briefly reports our attempts to stabilize the flight dynamics of Northeastern's Aerobat platform \cite{sihite_computational_2020} (Fig.~\ref{fig:cover}). 
This tailless platform can dynamically adjust its wing platform configurations during gaitcycles adding to its efficiency and agility. That said, Aerobat's morphing wings add to the inherent complexity of flight control and stabilization. This work continues our past efforts to control Aerobat's flight dynamics.

Flapping robots are not new. A plethora of examples is mainly dominated by insect-style design paradigms that are passively stable. Passive flight stabilization has a long history in flapping wing flight. Some systems device a tail \cite{de_croon_design_2009, rosen_development_2016, wissa2015free, chang2020soft, send2012artificial}, or design for a low center of mass to achieve open loop stability \cite{phan_kubeetle-s_2019, ma_controlled_2013}. However, tailless flapping flight is relatively new \cite{ramezani_biomimetic_2017}. 
Its advent coincided with the introduction of small sensors and powerful computational resources that these aerial robots could carry for active flight control. 

Aerobat, in addition to being tailless, unlike \cite{chukewad_robofly_2021, de_croon_design_2009, wissa2015free, chang2020soft, send2012artificial, ma_controlled_2013, phan_kubeetle-s_2019, rosen_development_2016}, is capable of significantly morphing its wing structure during each gait cycle. The robot has a weight of 40g when carrying a battery and a basic microcontroller, with an additional payload capacity of 20g, and a wingspan of 30 cm. Aerobat is powered by a 2-cell Lithium Polymer battery and is controlled onboard through a Raspberry Pi Zero 2w that also interacts with its camera and inertial measurement unit (IMU), used for autonomous localization through visual-inertial techniques.

The notion of embodiment in flapping wings, first was introduced and tested experimentally in \cite{ramezani_biomimetic_2017}. 
However, Aerobat possesses more complex embodied reconfiguration capabilities. Aerobat utilizes a computational structure called the \textit{Kinetic Sculpture} (KS) \cite{sihite_computational_2020, hoff_optimizing_2018}, which introduces computational resources for wing morphing. The KS is designed to actuate the robot's wings as it is split into two wing segments: the proximal and distal wings.

The morphing has energy efficiency benefits. The wing folding reduces the wing surface area and minimizes the negative lift during the upstroke, resulting in a more efficient flight. However, wing folding makes Aerobat very unstable. We have attempted outdoor flight using launchers to demonstrate Aerobat's thrust generation capabilities at high speeds \cite{siliconsynapse_lab_progress_2022}. 
However, maintaining a fixed position and orientation in hovering modes is ideally desirable because of the space constraints and diverse application of hovering aerial vehicles.

In this work, we aim to achieve stable hovering maneuvers. Since the actuation framework considered for Aerobat is being designed currently \cite{ramezani_aerobat_2022, sihite2021integrated, sihite_enforcing_2020, ramezani2020towards}, 
we are attempting other closed-loop operation options based on the addition of external position and orientation compensators to Aerobat using a guard as well. First, we cover the guard design. Next, we explain the model of Aerobat-guard considered in this paper. After, we describe our control approach, followed by results and concluding remarks. 

\begin{figure*}
\vspace{0.08in}
    \centering
    \includegraphics[width=1\linewidth]{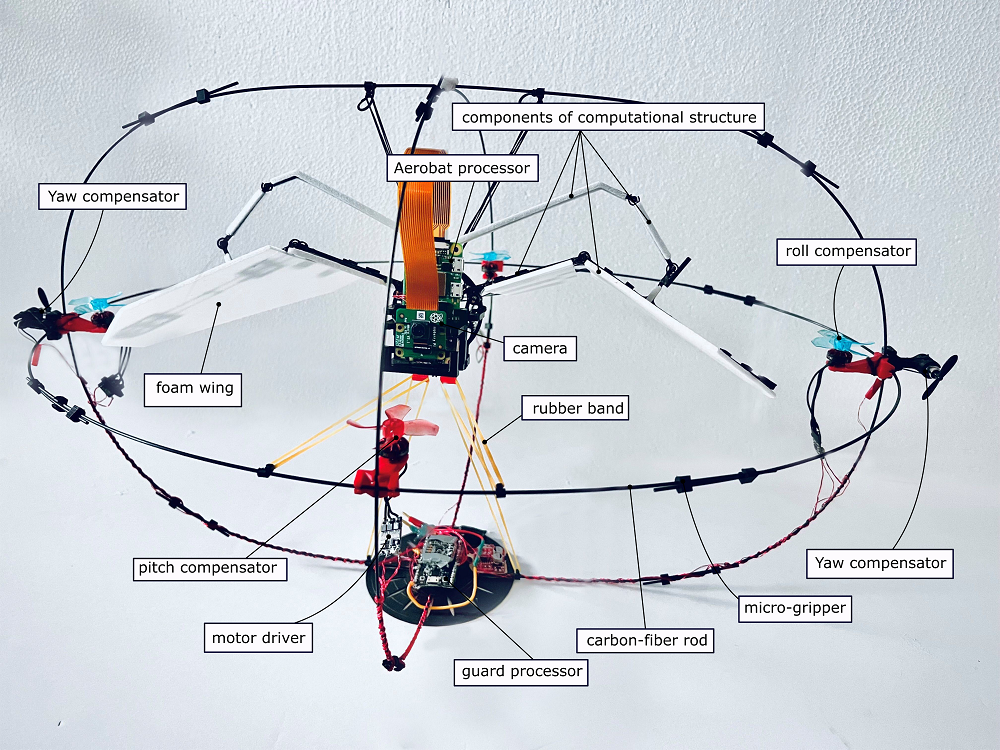}
    \caption{Illustrates Aerobat encapsulated inside a protecting guard.}
    \label{fig:guard}
\vspace{-0.08in}
\end{figure*}

\section{Overview of Guard Hardware}

To address the hovering challenge and prepare for the development of closed-loop control, we have built an active stabilization and passive protection system referred to here simply as \textit{The Guard} shown in Fig. \ref{fig:guard}. 

The guard is made up of a set of 11 lightweight 1.5mm carbon fiber rods, attached by PLA-3D printed pieces, and Aerobat is suspended at the center of the guard through four elastic rubber bands. The deformable elastic structure is designed to provide Aerobat with all-around protection in the event of a crash or collision with the environment. To actively stabilize Aerobat, the guard is also equipped with six small DC motors that stabilize the roll, pitch, yaw, and x-y-z positions of the robot. 

These thrusters can carry their own weight, thus nullifying the effect of the added weight of the guard and its electronics; however, they are not powerful to independently lift the system, that is, the lift force generated by Aerobat is required for hovering. To stabilize the yaw dynamics, the guard is equipped with two small and lightweight motors located at the ends of the long axis of the guard (see Fig.~\ref{fig:guard}). Due to the long arms, these motors need not be very powerful and can stabilize the roll, pitch, and yaw angles while adding minimal weight. 

The active stabilization offered by the guard allows Aerobat to test its thrust generation capabilities and, in small increments, exert control of the position and heading of the combined system. Over time, as Aerobat's controls become more stable and reliable through tuning and developing more accurate models, we will reduce the effort the guard applies on active stabilization and let Aerobat take full control of its flight. 

The guard is equipped with motion capture markers for closed-loop control and communicates wirelessly with an Optitrack motion capture system. It possesses an onboard micro-controller, ESP32. The ESP32 serves as the primary flight controller for the guard. It features a dual-core Tensilica LX6 microprocessor with clock speeds up to 240 MHz and integrated WiFi connectivity, which is used to transmit data from OptiTrack. The ESP32 controls the six electronic speed controllers. An IMU, ICM-20948, is utilized to estimate the guard's orientation in addition to the OptiTrack position and orientation data.


\section{Modeling}
We consider the dynamics of the guard and Aerobat as a multi-agent platform that interact. We assume the guard does not know Aerobat's states. 

\begin{figure*}
\vspace{0.08in}
    \centering
    \includegraphics[width=0.8\linewidth]{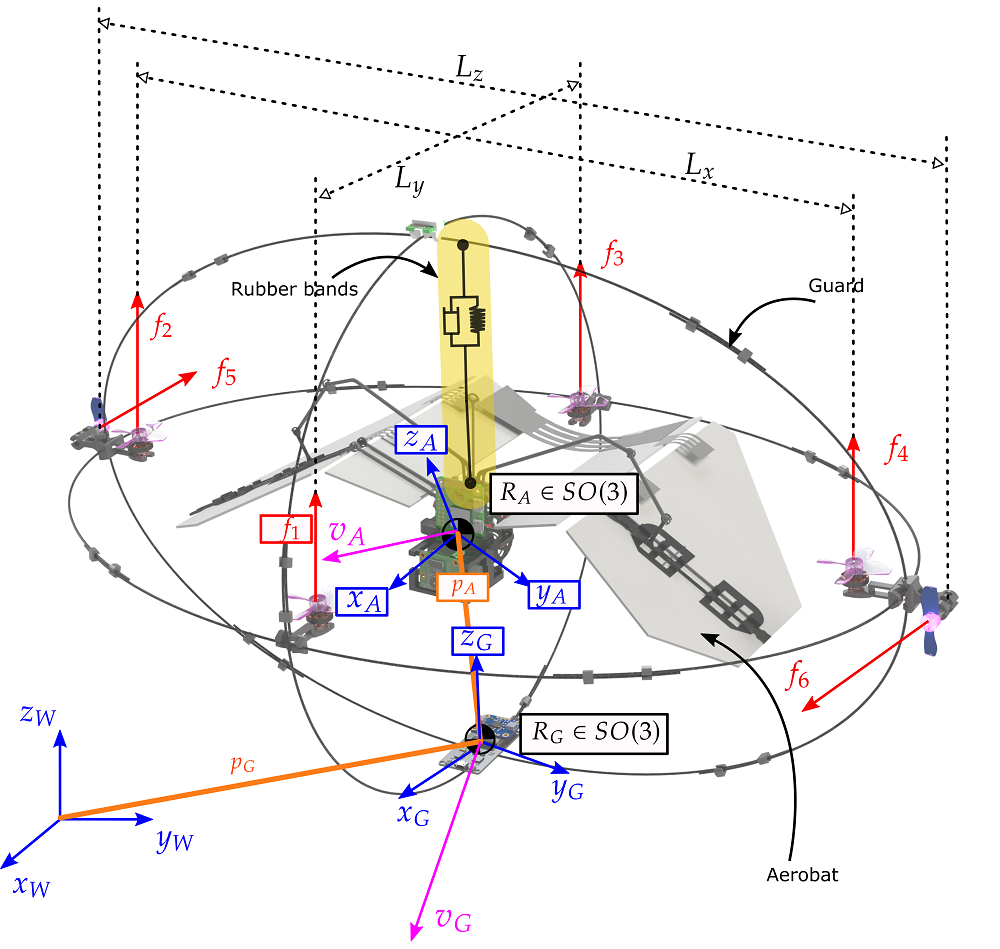}
    \caption{Illustrates Aerobat-guard free-body-diagram.}
    \label{fig:guard-fbd}
\vspace{-0.08in}
\end{figure*}

\begin{figure}
\vspace{0.08in}
    \centering
    \includegraphics[width=0.8\linewidth]{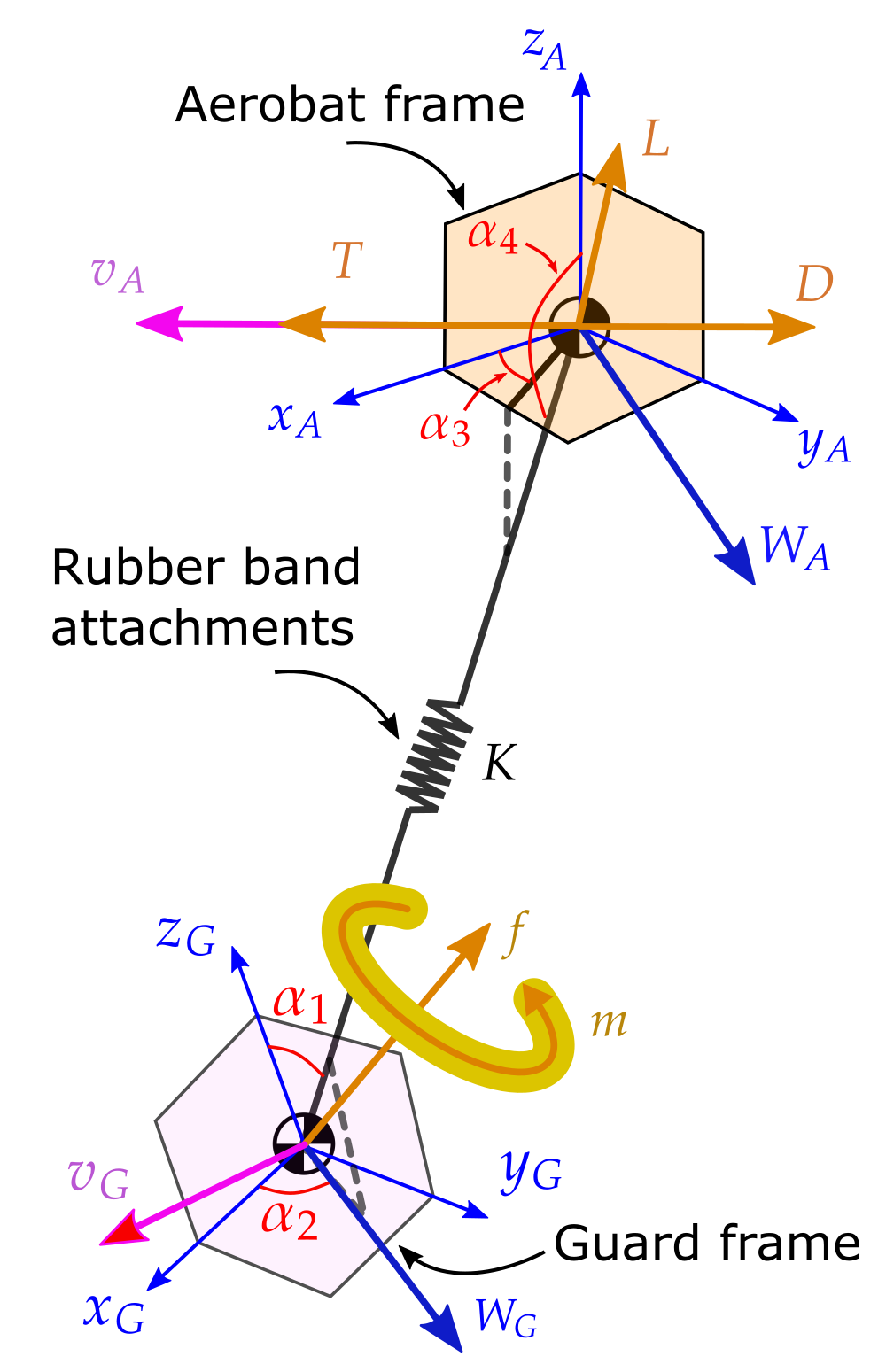}
    \caption{Shows the reduced-order-model (ROM) considered to describe Aerobat's bounding flight. $T$, $L$, and $D$ denote the thrust, lift, and drag forces that are not observable to the guard. Instead, they are added as an extended state ($x_3$ in Eq.~\ref{eq:observer}) to the guard dynamics and its estimated value used to control guard-Aerobat platform.}
    \label{fig:roms}
\vspace{-0.08in}
\end{figure}

We use a reduced-order model (ROM) (see Fig.~\ref{fig:roms}) of the guard-Aerobat platform for control design. The ROM has a total 14 DoFs, including 6 DoFs position and orientations of the guard (fully observable) and a total of 8 DoFs in Aerobat, including the position (3 DoFs), orientations (3 DoFs) relative to the guard, and wing joint angles (2 DoFs due to the symmetry in wings). This model is described with $q$, the configuration variable vector. 

Consider the guard's position $p_G$, orientation (Euler angles) $q_G=[q_x,q_y,q_z]^\top$, body-frame angular velocity vector $\omega_G$, mass $m_G$, and inertia $J_G$. The body-frame forces $f_i$ and moments $m_i$ acting on the guard are:
\begin{equation}
\begin{aligned}
& f=f_1+f_2+f_3+f_4+f_5+f_6+f_e \\
& m_x=L_x\left(f_4-f_2\right)+m_{e,x} \\
& m_y=L_y\left(f_3-f_1\right)+m_{e,y} \\
& m_z=L_z\left(f_6-f_5\right)+m_{e,z}
\end{aligned}
    \label{eq:body-force-moment}
\end{equation}
\noindent where $f_i,~~i=1-6$ are roll, pitch, and yaw compensators as shown in Fig.~\ref{fig:guard-fbd}. $L_i$ is shown in Fig.~\ref{fig:guard-fbd}. In Eq.~\ref{eq:body-force-moment}, $f_e$ denotes the rubber bands' pretension forces. In Eq.~\ref{eq:body-force-moment}, we assume the lift, drag, and thrust forces introduced by the Aerobat have impinged on the guard through the rubber bands' tension forces. While this assumption can be incorrect due to the complex wake interactions between Aerobat's wings and the guard propellers, the assumption permits the isolated modeling of these two systems using standard tools as follows. 

The body forces are mapped to the inertial-frame force by
\begin{equation}
\left[\begin{array}{c}
F_x \\
F_y \\
F_z
\end{array}\right]=R_G^0(q_G)\left[\begin{array}{c}
0 \\
0 \\
f
\end{array}\right]
\end{equation}
\noindent This force is described in the world frame using $R_G^0$. The general equations of motion of the guard while interacting with Aerobat are given:
\begin{equation}
\Sigma_{Guard}:\left\{\begin{array}{l}
\ddot p_G= -g[0,0,1]^\top+\frac{1}{m_G}F\\
\dot R_G^0= R_G^0 \hat\omega_G \\
J_G \dot\omega_G+\omega_G \times J_G \omega_G =m 
\end{array}\right.
    \label{eq:guard-modell}
\end{equation}
\noindent where $g=9.8~ms^{-2}$. 
The equations of motion of Aerobat are given by
\begin{equation}
\Sigma_{Aerobat}:\left\{\begin{array}{l}
[\ddot p_A^\top,\ddot q_A^\top]^\top = -D_u^{-1}\Big( D_{ua} \dot a(t)-H_u+J^\top y\Big)\vspace{0.1in}\\
\dot \xi=\Pi_1(\xi)\xi+\Pi_2(\xi)a(t) \vspace{0.1in}\\
y=\Pi_3(\xi)\xi+\Pi_4(\xi)a(t)
\end{array}\right.
    \label{eq:aerobat-full-dynamics}
\end{equation}
\noindent where $\Pi_i$ and $\xi$ are the aerodynamic model parameters and fluid state vector. $y$ is the output of the aerodynamic model, that is, the aerodynamic force. $J$ is the Jacobian matrix. $a(t)=[\dots a_i(t)\dots]^\top$ denotes wings' joint trajectories. $p_A$ and $q_A$ denote the position and orientation of Aerobat with respect to the guard and are given by
\begin{equation}
    \begin{aligned}
        p_A = R_A^G(q_A)[0,0,1]^\top\\
        q_A = [\alpha_3,\alpha_4]^\top
    \end{aligned}
\end{equation}
\noindent Please see Fig.~\ref{fig:guard-fbd} for more information about $\alpha_i$. In Eq.~\ref{eq:aerobat-full-dynamics}, $D_u$, $D_{ua}$, and $H_u$ are the block matrices from partitioning Aerobat's full-dynamics. For partitioning, the full-dynamics inertia matrix $D(q)$, Coriolis matrix $C(q,\dot q)$ and conservative potential forces (gravity and rubber bands) $G(q)$ corresponding to the underactuated (position and orientation) $q_u=[p_A^\top,q_A^\top]^\top$ and actuated (the wings joints) $a(t)$ are separated. 

The force and moments $f_e$ and $m_e$ in Eq.~\ref{eq:body-force-moment} are obtained as elastic conservative forces. To do this, the total potential energy $V(q)$ is used to obtain $H=C(q,\dot q)\dot q + G(q)$ in Aerobat's model.  $V(q)$ is given given by
\begin{equation}
    V(q)=\frac{1}{2}K(p_G-p_A)^\top(p_G-p_A) \\+ m_Ag\Big(p_{G,z}+R_A^0p_{A,z}\Big)
    \label{eq:total-potential-energy}
\end{equation}
\noindent where $K$ denotes the rubber bands' elastic coefficient and $m_A$ is Aerobat's total mass. Note that in Eq.~\ref{eq:aerobat-full-dynamics}, we model aerobat-fluidic environment interactions using our experimentally tested unsteady-quasi-static models that are evaluated at discrete strips along the wing span. The output from this aerodynamic model $y=[y1,\dots,y_m]^\top$ where $y_i$ is the external aerodynamic force at i-th strip is governed by the state-space model $\dot \xi$. We briefly cover this model; however, the reader is referred to \cite{sihite2022unsteady} for more details on the model derivations. 

\subsection{Brief Overview of Unsteady Aerodynamic State-Space Model $\dot \xi=A(\xi)\xi+B(\xi)a(t)$}

We superimpose horseshoe vortices on and behind the wing blade elements to calculate lift and drag forces. Consider the time-varying circulation value at $s_i$, the location of the $i$-th element on the wing, denoted by $\Gamma_i(t)$. The circulation can be parameterized by truncated Fourier series of $n$ coefficients. $\Gamma_i$ is given by 
\begin{equation}
\begin{aligned}
    \Gamma_i(t) = a^\top(t)
    \begin{bmatrix}
    \sin(\theta_i)\\
    \vdots\\
    \sin(n\theta_i)\\
    \end{bmatrix},
\end{aligned}
    \label{eq:circulation}
\end{equation}
\noindent where $a=[a_1,\dots,a_n]^\top$ are the Fourier coefficients and $\theta_i=\arccos(\frac{s_i}{l})$ ($l$ is the wingspan size). From Prandtl's lifting line theory, additional circulation-induced kinematics denoted by $y_{\Gamma}$ are considered on all of the points $p_i$. These circulation-induced kinematics are given by
\begin{equation}
\begin{aligned}
    y_{\Gamma} = 
    \begin{bmatrix}
    1&\frac{\sin{2\theta_1}}{\sin{\theta_1}}&\dots&\frac{\sin{n\theta_1}}{\sin{\theta_1}}\\
    1&\frac{\sin{2\theta_2}}{\sin{\theta_2}}&\dots&\frac{\sin{n\theta_2}}{\sin{\theta_2}}\\
    \vdots\\
    1&\frac{\sin{2\theta_n}}{\sin{\theta_n}}&\dots&\frac{\sin{n\theta_n}}{\sin{\theta_n}}\\
    \end{bmatrix}a(t)
    \label{eq:induced-kin}
\end{aligned}
\end{equation}
Now, we utilize a Wagner function $\Phi(\tau) = \Sigma_{k=1}^2\psi_k \exp{(-\frac{\epsilon_k}{c_i} \tau)}$, where $\psi_k$, $\epsilon_k$ are some scalar coefficients and $\tau$ is a scaled time to compute the aerodynamic force coefficient response $\beta_{i}$ associated with the i-th blade element. The kinematics of i-th element using Eqs.~\ref{eq:aerobat-full-dynamics} and \ref{eq:induced-kin} are given by 
\begin{equation}
y'_{1,i} = y_{1,i} + y_{\Gamma,i}.    
\end{equation}
\noindent Following Duhamel's integral rule, the response is obtained by the convolution integral given by
\begin{equation}
    \beta_i = y'_{1,i}\Phi_0 + \int_0^t\frac{\partial \Phi(t-\tau)}{\partial \tau}y'_{1,i}d\tau \\
\label{eq:lift_coeff_wagner}
\end{equation}
\noindent where $\Phi_0=\Phi(0)$. We perform integration by part to eliminate $\frac{\partial \Phi}{\partial \tau}$, substitute the Wagner function given above in \eqref{eq:lift_coeff_wagner}, and employ the change of variable given by $z_{k,i} (t) = \int_{0}^{t} \exp{(-\frac{\epsilon_k}{c_i}(t-\tau))} y'_{1,i} d\tau$, where $k\in\{1,2\}$, to obtain a new expression for $\beta_i$ based on $z_{k,i}$  
\begin{equation}
\beta_i =
y'_{1,i}\Phi_0+
\begin{bmatrix}
\psi_1\frac{\epsilon_1}{c_i}&\psi_2\frac{\epsilon_2}{c_i}
\end{bmatrix}
\begin{bmatrix}
z_{1,i}\\
z_{2,i}
\end{bmatrix}
\label{eq:?1}
\end{equation}
The variables $a$, $z_{1,i}$, and $z_{2,i}$ are used towards obtaining a state-space realization that can be marched forward in time. Using the Leibniz integral rule for differentiation under the integral sign, unsteady Kutta–Joukowski results $\beta_i=\frac{\Gamma_i}{c_i}+\frac{d\Gamma_i}{dt}$, and \eqref{eq:circulation}, the model that describes the time evolution of the aerodynamic states is obtained
\begin{equation}
\Sigma_{Aero,i}:\left\{
\begin{aligned}
    A_i\dot{a} &= -B_ia + C_iZ_i  + \Phi_0y'_{1,i}\\
    \dot{Z}_i &= D_iZ_i + E_iy'_{1,i}\\
\end{aligned}
\right.
    \label{eq:aerodynamic-ss-model}
\end{equation}
\noindent where $Z_i$, $A_i$, $B_i$, $C_i$, $D_i$, and $E_i$ are the state variable and matrices corresponding to the i-th blade element. They are given by
\begin{equation}
\begin{aligned}
    Z_i &= 
    \begin{bmatrix}
    z_{1,i} & z_{2,i}
    \end{bmatrix}^\top,\\
    A_i &= 
    \begin{bmatrix}
    \sin{\theta_i} & \sin{2\theta_i} & \dots & \sin{n\theta_i} 
    \end{bmatrix},\\
    B_i &= A_i/c_i,\\
    C_i &= 
    \begin{bmatrix}
    \frac{\psi_1\epsilon_1}{c_i} & \frac{\psi_2\epsilon_2}{c_i}
    \end{bmatrix},\\
    D_i &= 
    \begin{bmatrix}
    \frac{-2\epsilon_1}{c_i} & 0\\
    0 & \frac{-2\epsilon_2}{c_i}
    \end{bmatrix},\\
    E_i &=
    \begin{bmatrix}
    2-\exp{\frac{\epsilon_1 t}{c_i}} & 2-\exp{\frac{\epsilon_2 t}{c_i}}
    \end{bmatrix}^\top.
\end{aligned}
    \label{eq:?2}
\end{equation}
\noindent We define the unified aerodynamic state vector, used to describe the state space of \eqref{eq:aerobat-full-dynamics}, as following $\xi = [a^\top, Z^\top]^\top$, where $Z=[Z^\top_1,\dots,Z^\top_n]^\top$.

\section{Control}

The full dynamics of the guard with Aerobat sitting inside are given by
\begin{equation}
\Sigma_{FullDyn}\left\{\begin{array}{l}
\dot{x}_1=x_2 \\
\dot{x}_2=g_1+g_2u+g_3x_3 \\
\dot{x}_3=G(t) \\
z=x_1
\end{array}\right.
    \label{eq:extended-state-model}
\end{equation}
\noindent where $x_1=[p_G^\top,q_G^\top]^\top$, the nonlinear terms $g_i$ are given by Eqs.~\ref{eq:guard-modell} and \ref{eq:aerobat-full-dynamics}, $u=[\dots f_i \dots]$ from Eq.~\ref{eq:body-force-moment}, and $x_3=y$ from Eq.~\ref{eq:aerobat-full-dynamics}. As it can be seen, the model is extended with another state $x_3$ because $y$ (aerodynamic force) can be written in the state-space form given by Eq.~\ref{eq:aerobat-full-dynamics}. 

Now, the control of Eq.~\ref{eq:extended-state-model} is considered here, assuming that $u$ is calculated only based on $z=x_1$ observations. The time-varying term $G(t)$ (the dynamics of $y$) is highly nonlinear; however, we have an efficient model of $G(t)$ \cite{sihite2022unsteady}. We obtained an aerodynamic model in \cite{sihite2022unsteady} that closely predicts the external aerodynamic forces impinged on Aerobat. 

Using this model, we establish a state observer for $x_3$ to augment the feedback $u=Kx_2$ (where $K$ is the control gain) so that $\dot x_2$ remains bounded and stable. Consider the following definition of estimated states $\hat x_i$ from \cite{sihite2022unsteady}:
\begin{equation}
\begin{aligned}
& \hat{x}_1=\hat{x}_2-\beta_1\left(\hat{x}_1-x_1\right) \\
& \hat{x}_2=g_1+g_2 u+g_3\hat{x}_3-\beta_2\left(\hat{x}_1-x_1\right) \\
& \hat{x}_3=-\beta_3\left(\hat{x}_1-x_1\right)
\end{aligned}
    \label{eq:estimated-states}
\end{equation}
\noindent where $\beta_i$ is the observer gains. Now, we define the error $e_i=\hat x_i- x_1$ for $i=1,2,3$. The following observer model is found
\begin{equation}
\left[\begin{array}{c}
\dot e_1 \\
\dot e_2\\
\dot e_3
\end{array}\right]=\left[\begin{array}{ccc}
-\beta_1 & I & 0 \\
-\beta_2 & 0 & g_3 \\
-\beta_3 & 0 & 0
\end{array}\right]\left[\begin{array}{c}
e_1 \\
e_2 \\
e_3
\end{array}\right]+\left[\begin{array}{c}
0 \\
0 \\
-I
\end{array}\right] G(t)
    \label{eq:observer}
\end{equation}
\noindent The gains $\beta_i$ for the observer given by Eq.~\ref{eq:observer} can be obtained if upper bounds for $G$ and $g_2$ can be assumed. We have extensively studied $g_1$, $g_2$ and $g_3$ terms in Aerobat's model in past and ongoing efforts. Based on the bounds for $\|g_1\|$, $\|g_2\|$, and $\|g_3\|$, we tuned the observer. The controller used for the bounding flight is given by
\begin{equation}
u = g_2^{-1}\Big(u_0 -g_1 - g_3\hat x_3\Big) 
    \label{eq:controller}
\end{equation}
\noindent where $u_0=Kx_2$.

\section{Results}

\begin{figure*}
\vspace{0.08in}
    \centering
    \includegraphics[width=1\linewidth]{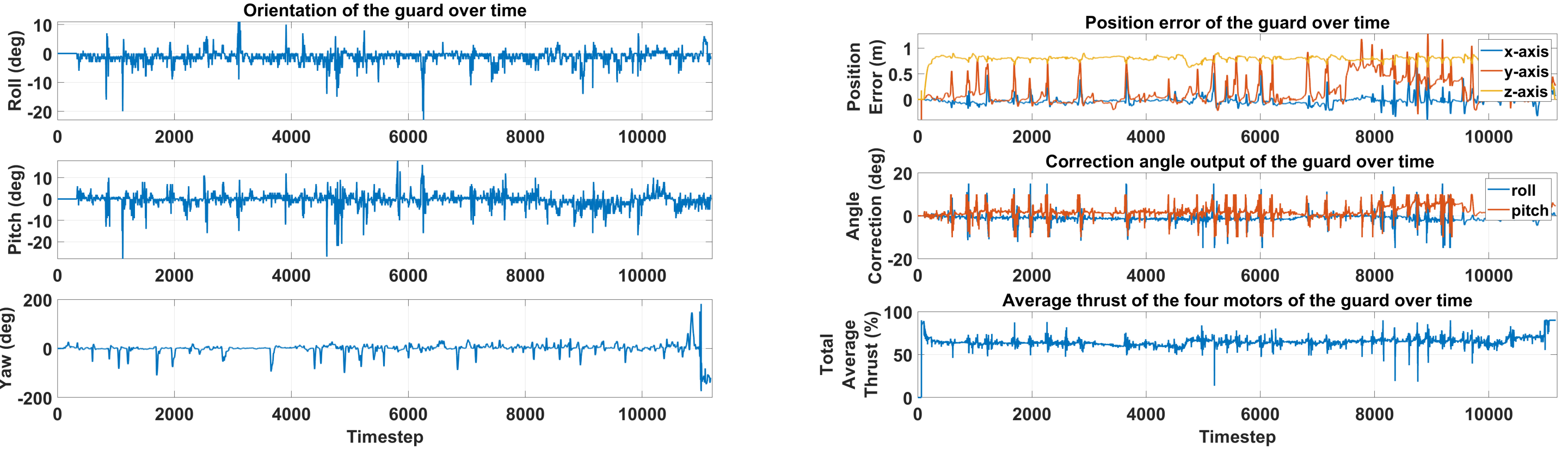}
    \caption{Shows the tracking performance of the controller based on estimated generalized inertial $g_1^{-1}f$ and aerodynamics $g_1^{-1}g_2\hat x_3$ to maintain a fixed position and orientation of the guard.}
    \label{fig:optitrack}
\vspace{-0.08in}
\end{figure*}

\begin{figure}
\vspace{0.08in}
    \centering
    \includegraphics[width=1\linewidth]{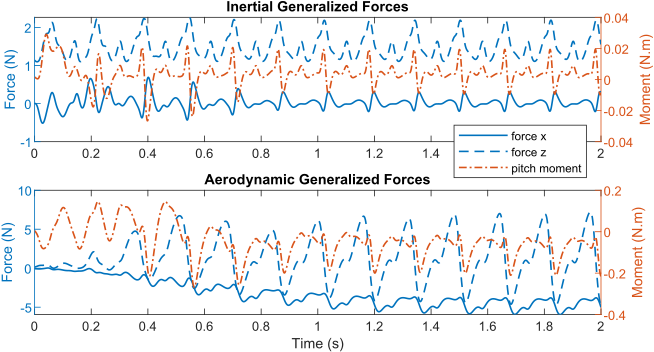}
    \caption{Shows estimated generalized inertial $g_2^{-1}g_1$ and aerodynamics $g_2^{-1}g_3\hat x_3$ contributions based on knowledge on the boundedness of \textbf{$\|g_2\|$} and \textbf{$\|G(t)\|$}.}
    \label{fig:gen_force}
\vspace{-0.08in}
\end{figure}

The calculation of the control command  $u=[f_1,\dots,f_6]^\top$ was achieved based on estimated values of the extended state $x_3$ in hovering inside our lab. $x_2$, which embodies the guard's world position and orientations and velocities, were measured using our OptiTrack system and the onboard inertial measurement unit. Our unsteady aerodynamic model reported in \cite{sihite2022unsteady} was used to identify the bounds on $G(t)$ during hovering. Based on unsteady aerodynamic model results, for hovering, $\|G(t)\|$ was selected. The estimated values $g_1^{-1}g_2\hat x_3$ for hovering denoted as generalized aerodynamic force contributions are shown in Fig.~\ref{fig:gen_force}. The estimated values for $g_1^{-1}f$ denoted as generalized inertial dynamics contributions and shown in Fig.~\ref{fig:gen_force} were used to complete the computation of $u$. The performance of the controller in stabilizing the roll, pitch, yaw, x-y-z positions are shown in Figs.~\ref{fig:optitrack}. 

\section{Concluding Remarks}

Aerobat, in addition to being tailless, possesses morphing wings that add to the inherent complexity of flight control. In this work, we covered our efforts to stabilize the flight dynamics of Aerobat. We employed a guard design with manifold small thrusters to stabilize Aerobat's position and orientation in hovering. We developed a dynamic model of Aerobat and guard interacting with each other. For flight control purposes, we assumed the guard cannot observe Aeroat's states. Then, we proposed an observer design to estimate the unknown states of the Aerobat-guard dynamics, which were used for closed-loop hovering control. We reported culminating experimental results that showcase the effectiveness of our approach.

\printbibliography

\end{document}